\documentclass[sigconf]{acmart}
\usepackage{algorithmic}
\usepackage{graphicx}
\usepackage{textcomp}
\usepackage{xcolor}
\usepackage{dblfloatfix}
\def\BibTeX{{\rm B\kern-.05em{\sc i\kern-.025em b}\kern-.08em
    T\kern-.1667em\lower.7ex\hbox{E}\kern-.125emX}}
    
\usepackage{enumitem}

\usepackage{accents}

\usepackage{soul,color}

\newcommand{\name}{{\tt STC-GEF}}

\setcopyright{none}
\renewcommand\footnotetextcopyrightpermission[1]{}
\usepackage{multirow}

\acmConference[UrbComp '22]{UrbComp '22: The 11th SIGKDD International Workshop on Urban Computing}{August 15, 2022}{Washington, DC}

\author{Mahan Tabatabaie  \quad James Maniscalco \quad Connor Lynch \quad Suining He
}
\affiliation{%
   \institution{University of Connecticut}
   \city{Storrs}
   \state{CT}
   \country{USA}}
\email{{mahan.tabatabaie, james.maniscalco, connor.lynch, suining.he}@uconn.edu}

\renewcommand{\shortauthors}{M. Tabatabaie, J. Maniscalo, C. Lynch, and S. He}

\begin{document}

\title{Towards Spatio-Temporal Cross-Platform Graph Embedding Fusion for Urban Traffic Flow Prediction}

\begin{abstract}
In this paper, we have 
proposed \name{}, a novel \textbf{S}patio-\textbf{T}emporal \textbf{C}ross-platform 
\textbf{G}raph \textbf{E}mbedding \textbf{F}usion approach
for the urban traffic flow prediction. 
We have designed a \textit{spatial embedding} module based on graph convolutional networks (GCN)
to extract the complex \textit{spatial} features within the traffic flow data. Furthermore, 
to capture the \textit{temporal} dependencies between the traffic flow data from various time intervals, 
we have designed a \textit{temporal embedding module} 
based on recurrent neural networks. 
 Based on the observations that 
 different transportation platforms' trip data (\textit{e.g.,} taxis, Uber, and Lyft) 
 can be correlated, we have designed an effective fusion mechanism that combines 
 the trip data from different transportation platforms and further uses them for 
 cross-platform traffic flow prediction (\textit{e.g.,} integrating taxis and ride-sharing platforms for 
 taxi traffic flow prediction).
 We have conducted extensive real-world experimental studies based on real-world trip data of
 yellow taxis and ride-sharing (Lyft) from New York City (NYC), 
 and validated
 the accuracy and effectiveness of \name{} in fusing different transportation platform data and predicting 
 traffic flows.

\end{abstract}


\maketitle

\section{Introduction}

Urban traffic prediction for transportation platforms such as taxis, 
ride-sharing, as well as other public/private vehicles has attracted much attention recently due to the important social and business values~\cite{traffic-news,alexander2015assessing,10.1145/3534603,10.1145/3474717.3483911}. With urban traffic prediction and emerging big mobility data, various ubiquitous 
and urban computing applications have been enabled or strengthened, including vehicle routing~\cite{santos2018context},
event monitoring~\cite{kaiser2017advances}, and autonomous driving.
However, two major technical challenges remain when deploying 
a large-scale data-driven traffic prediction system:
\begin{itemize}
     \item 
     \textbf{Dynamic and complex mobility 
     activity/usage patterns in different regions:} 
     To predict the traffic flow in a future time interval, 
     one may use the historical traffic flow data to calculate spatial and temporal features. 
     However, there exist various types of traffic patterns in the historical data and designing a model 
     to effectively capture the most critical and related ones based on different conditions 
     is highly challenging. 
    \item \textbf{Absence of cross-platform mobility data fusion:}
    The traffic flow is highly correlated to
    different transportation platforms such as bike sharing or ride-sharing systems 
    (\textit{e.g.,} Lyft and Uber). Therefore, to have an accurate prediction of the traffic flow, 
    such correlation must be identified and effectively taken into account, which is a challenging task.
\end{itemize}

To address the above challenges, we 
propose \name{}, the novel \textbf{S}patio-\textbf{T}emporal \textbf{C}ross-platform \textbf{G}raph \textbf{E}mbedding \textbf{F}usion approach
for the urban taxi flow prediction. 
In this prototype study, we have made the following three major contributions:
\begin{enumerate}
 \item \textbf{Spatial and Temporal Graph Embedding Learning:}
 To extract the complex spatial features within traffic flow data, we propose a spatial embedding module based on graph convolutional networks (GCN). Additionally, to capture the temporal dependencies between the traffic flow data from various time intervals, we leverage a temporal embedding module based on recurrent neural networks. 
 
 \item \textbf{Cross-platform Mobility Data Fusion:} 
 Based on our data analysis, we have identified that different transportation 
 platforms' trip data (\textit{e.g.,} taxis, Uber, and Lyft) may be highly correlated. 
 Therefore, we have proposed an effective fusion mechanism that combines the 
 trip data from different transportation platforms and further uses them for cross-platform traffic flow prediction.
 In particular, in the preliminary studies, we leverage historical taxi and Lyft trip data to predict the future traffic flow of the taxis in NYC.
 
 \item \textbf{Extensive Experimental Evaluations:} We have performed data analytics and experimental studies on two public real-world datasets to evaluate the effectiveness of our \name{}. 
 Specifically, we have studied a total of 4,464,090 yellow taxi and Lyft trip records provided by the NYC's open data program~\cite{nyc-open-data, tlc-data} and shown that our proposed framework outperforms the baseline methods for traffic flow prediction.
\end{enumerate}

The rest of our paper is organized as follows. 
We first review some of the previous studies for traffic flow prediction in Sec.~\ref{sec:related-works}. 
Then, we present an overview of our framework in Sec.~\ref{sec:overview} followed by dataset details, problem formulation, and framework designs in Sec.~\ref{sec:dataset-problem}. 
We demonstrate our experimental studies in Sec.~\ref{sec:experiments}, and finally
conclude our work in Sec.~\ref{sec:conclusion}.

\section{Related Works}
\label{sec:related-works}

We briefly go through the related studies as follows. 
Various traditional time series forecasting methods have been explored for traffic flow prediction.
Chen~\textit{et al.}~\cite{chen2011short} leveraged the auto-regressive integrated moving average for traffic prediction. On the other hand, Kumar~\textit{et al.}~\cite{kumar2017traffic} designed a Kalman Filter to forecast the traffic flow. Dong~\textit{et al.}~\cite{dong2018short} used the Gradient Boosting Decision Tree algorithm to perform short-term traffic flow prediction.

With the recent advances of big mobility data~\cite{yang2021spatio} and deep learning, 
many recent studies have proposed various network architectures based on convolutional neural networks (CNN), 
long short-term memory (LSTM) or residual neural networks (ResNet~\cite{ResNet-Original}) to perform traffic flow prediction. 
Zhang \textit{et al.} \cite{zhang2017deep} proposed DeepST, 
a network based on CNN and ResNet modules to predict the crowd flow. 
Yang~\textit{et al.}~\cite{yang2021spatio} proposed a method based on graph neural networks and attention mechanism to predict the crowd inflow/outflow of different buildings. Zheng \textit{et al.} \cite{zheng2020gman} converted 
the road network to a weighted graph and leveraged graph neural networks with attention mechanisms for traffic flow prediction. 
Likewise, Shi~\textit{et al.}~\cite{shi2020spatial} proposed a graph neural network with a novel attention mechanism that considers both long-term and short-term periodical dependencies. 
Pan~\textit{et al.}~\cite{pan2020spatio} proposed a framework based on graph neural networks, which is trained based on meta learning to predict the traffic flow for multiple locations at the same time. 
Zhang~\textit{et al}.~\cite{zhang2018combining} combined the external factors such as weather conditions to better predict the traffic flow.  
Different from these prior studies, we propose a novel graph embedding approach that
fuses the different transportation platforms to further enhance the prediction accuracy.
We have designed a novel mobility data fusion approach and validated the performance with
real-world transportation platform data. 

\begin{figure}[tp]
    \centering
    \includegraphics[width = \columnwidth]{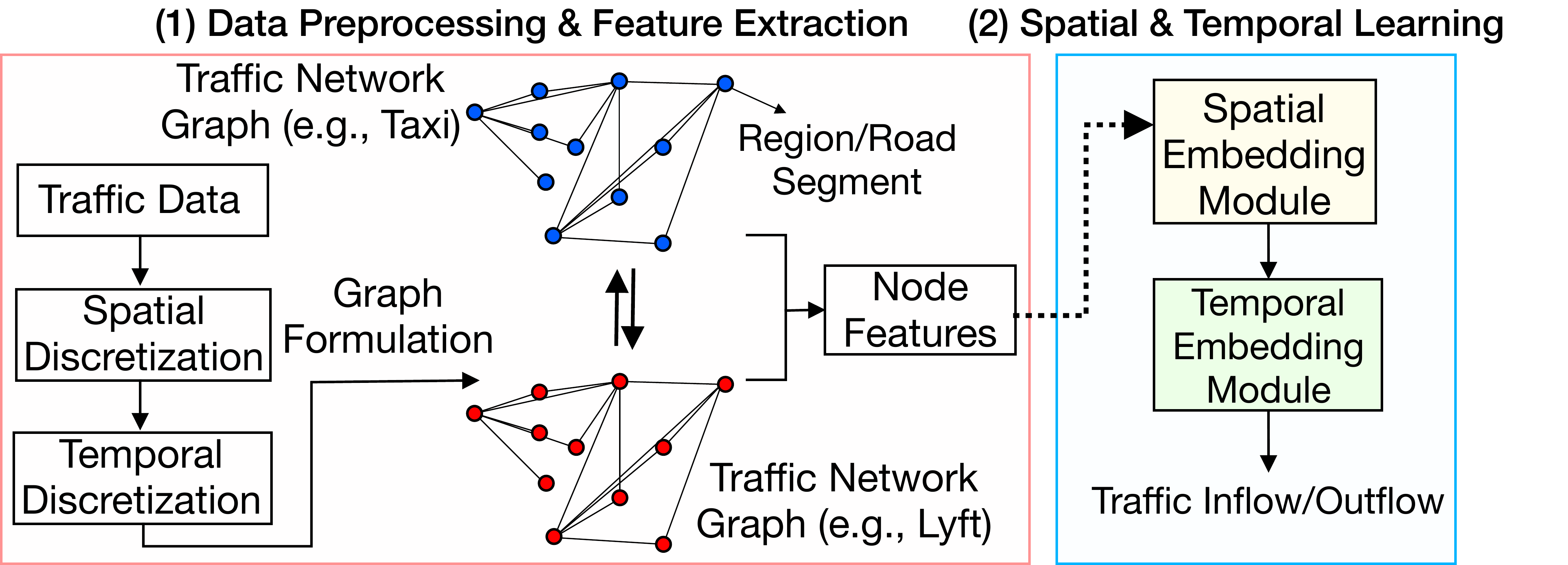}
    \caption{Overall information flow of \name{}.}
    \label{fig:overview}
\end{figure}

\section{Core Model Overview}
\label{sec:overview}
We present the overall workflow of our proposed framework in Fig.~\ref{fig:overview}, which consists of two main stages as follows:

\textbf{(1) Data Preprocessing and Feature Extraction:} In the preprocessing stage, we first perform spatial discretization. Specifically, we use the same taxi regions specified by the NYC open data program~\cite{nyc-open-data}, which are visualized in \cite{nyc-zones} to divide the target city into multiple regions. Moreover, each region is a polygon with its coordinates specified on the city map. Then, a unique number is assigned to each region, which is used to specify the pickup and drop-off location of the trips. Next, we perform temporal discretization by dividing each day into $P$ equal time intervals. Afterward, we convert the traffic data into multiple directed weighted graphs represented by adjacency matrices, in which the nodes represent the taxi regions and the edges are the traffic inflow/outflow between different regions. Finally, for each node, we calculate the sum of the traffic inflow and outflow from other regions and consider them as the \textit{node features}.

\textbf{(2) Spatial and Temporal Learning:} After the preprocessing and feature extraction stage, we feed the extracted node features and graph adjacency matrices to the \textit{spatial embedding module} that converts them into spatial embeddings. Then, the output of the spatial embedding module is further processed by the 
\textit{temporal embedding module} to produce the final spatial and temporal embedding of the input graphs, which is used to predict the traffic inflow/outflow for the future time interval.

\section{Datasets \& Problem Definition}
\label{sec:dataset-problem}

\subsection{Overview of Datasets}
Our datasets used in this paper are presented as follows.

\textbf{$\bullet$ Yellow Taxi Trip Data:}
We collected the yellow taxi trip data from NYC's open data program \cite{nyc-open-data} from $2021/01/01$ to $2021/01/31$ time period, which includes 1,369,765 taxi trips with the pickup and drop-off locations in a total of 265
taxi regions~\cite{nyc-zones}. 

\textbf{$\bullet$Lyft Trip Data:} Similarly, we collected the Lyft trip data \cite{tlc-data} from the
the time period of $2021/01/01$ to $2021/01/31$, which includes 3,094,325 Lyft trips with the pickup and drop-off locations in the same taxi regions~\cite{nyc-zones}.

\subsection{Model Design Motivations}
\begin{figure}[!ht]
\centerline{\includegraphics[width = \columnwidth]{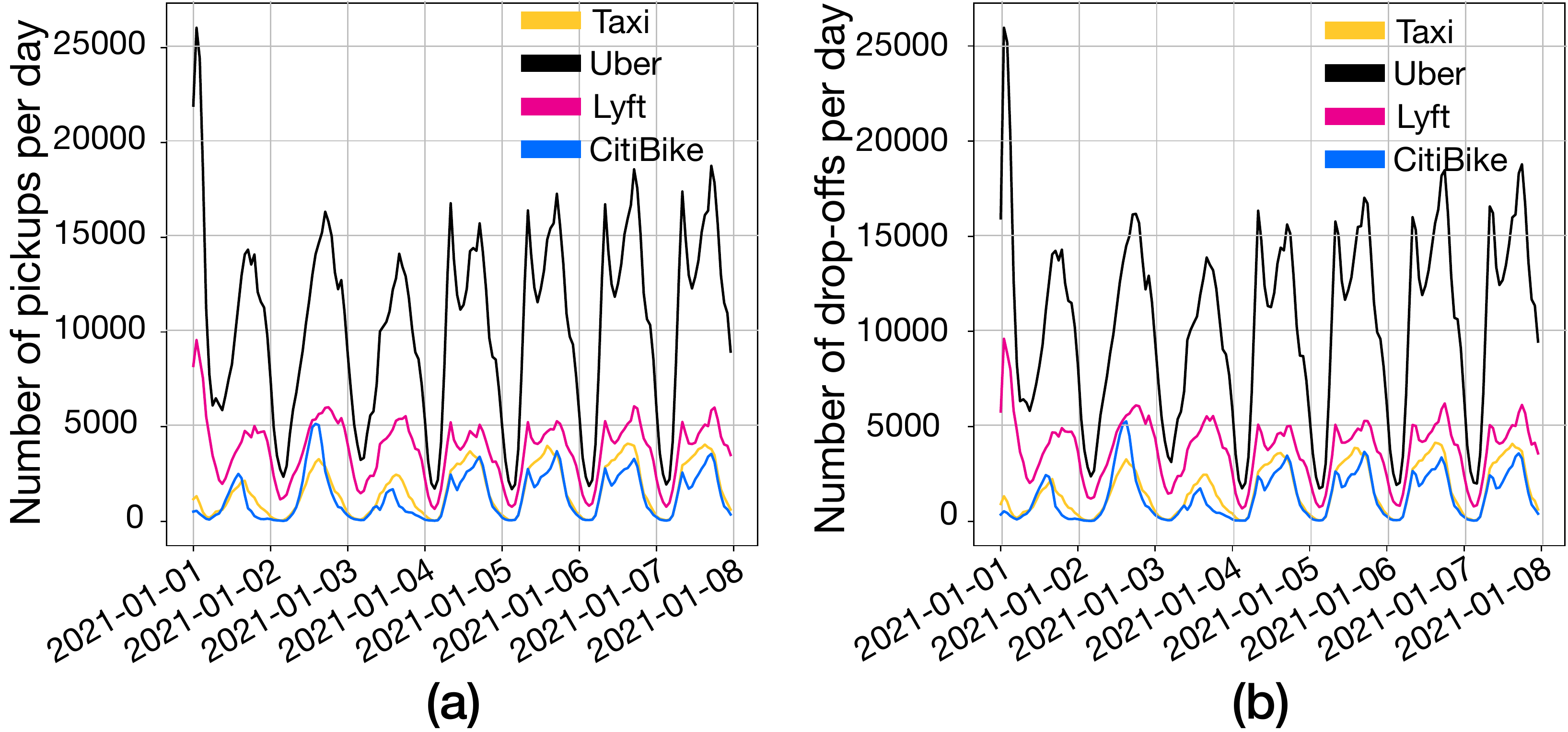}}
\caption{Comparison of pickup and drop-off data of different transportation platforms over a week that are accumulated 
every 3 hours. 
}
\label{fig:sample-data-week}
\end{figure}

To support our motivations, we illustrate the pickup and drop-off data of different transportation platforms (\textit{i.e.,} Taxi~\cite{nyc-open-data}, Citi Bike~\cite{bike-data}, Uber, and Lyft~\cite{tlc-data}) for a week in January 2021, which are accumulated for each day in Fig.~\ref{fig:sample-data-week}. 
From the figures, we can observe that different transportation platforms have similar and correlated pickup (Fig.~\ref{fig:sample-data-week}(a)) and drop-off (Fig.~\ref{fig:sample-data-week}(b)) trends. Therefore, we are motivated to combine different traffic flow data from various transportation platforms to improve our model's prediction accuracy.

\subsection{Problem Formulation}
As stated earlier, we formulate the traffic flow data as graphs to further use for future inflow/outflow predictions. Moreover, we consider each taxi region shown in \cite{nyc-zones} as a node and calculate the number of the taxi/Lyft trips between them as the weight of their edges and form the adjacency matrices. We denote the adjacency matrices created for the taxi trips on the $d$-th day and the $t$-th time interval by $\mathbf{G}^\text{Taxi}_{d,t} \in \mathbb{R}^{M\times M}$, where $M$ is the number of the taxi regions (265 in this study). Similarly, we denote the adjacency matrices based on the Lyft trips for the $d$-th day and the $t$-th time interval by $\mathbf{G}^\text{Lyft}_{d,t} \in \mathbb{R}^{M\times M}$.

Taking the taxi trips as an example, given the adjacency matrices of the $t$-th time interval in the $d$-th day, we calculate the sum of the inflow and outflow trips to each node (\textit{i.e.,} region) and consider them as two \textit{node features}. We denote the node features extracted from the taxi trips at $t$-th time interval in the $d$-th day by $\mathbf{N}^\text{Taxi}_{d,t} \in \mathbb{R}^{2\times M}$. Similarly, we denote the node features of the Lyft trips at the $t$-th time interval in the $d$-th day by $\mathbf{N}^\text{Lyft}_{d,t} \in \mathbb{R}^{2\times M}$, which are similarly calculated as mentioned above.

Given the historical adjacency matrices and node features of the taxi trip data for the $k$ time intervals prior to target time interval $T$ of the $d$-th day, $\{\mathbf{G}^\text{Taxi}_{d,T-k}, \dots, \mathbf{G}^\text{Taxi}_{d,T-1}\}$, $\{\mathbf{N}^\text{Taxi}_{d,T-k}, \dots, \mathbf{N}^\text{Taxi}_{d,T-1}\}$, and the historical Lyft adjacency matrices and node features of the Lyft trip data for the target time interval $T$-th of the $(d-1)$-th day, $\mathbf{G}^\text{Lyft}_{d-1,T}$, $\mathbf{N}^\text{Lyft}_{d-1,T}$, the goal of \name{} is to predict the traffic inflow and outflow of the taxi regions for the the $T$-th time interval of the $d$-th day, $\mathbf{N}^\text{Taxi}_{d,T}$. 
We note that since our target is to predict for a future $T$-th time interval of the $d$-th day, it would be impractical to assume the availability of the Lyft trip data for the exact same $T$-th time interval in the $d$-th day. Therefore, we consider the Lyft trip data of the same time period of a day before the target date, $(d-1)$, 
as we observed that two consecutive days may exhibit correlations
regarding their traffic inflow/outflow patterns (Fig.~\ref{fig:sample-data-week}).

\begin{figure}[tp]
    \centering
    \includegraphics[width =1.0\columnwidth]{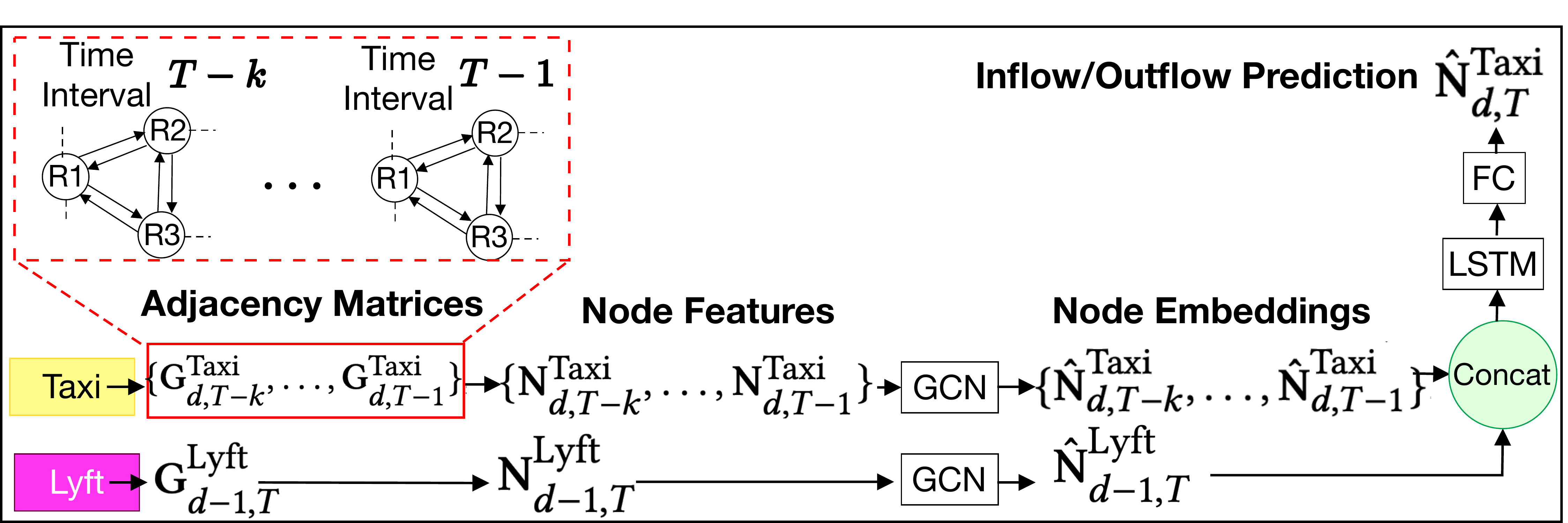}
    \caption{Information flow overview of \name{}.}
    \label{fig:core-overview}
\end{figure}

We present the overall architecture of \name{} in Fig.~\ref{fig:core-overview}. In particular, it first inputs the historical adjacency matrices and node features of the taxi trip data from the time interval $(T-k)$ to $(T-1)$ of the $d$-th day to the graph convolutional network (\texttt{GCN}) to produce the spatial node embeddings, $\{\mathbf{\hat{N}}^\text{Taxi}_{d,T-k}, \dots, \mathbf{\hat{N}}^\text{Taxi}_{d,T-1}\}$. Similarly, the historical adjacency matrices and node features of the Lyft trip data in the $T$-th time interval of the $(d-1)$-th day  are converted to spatial embeddings, $\mathbf{\hat{N}}^\text{Lyft}_{d-1,T}$. Next, the output spatial embedding is concatenated together and passed through the temporal embedding module to produce the spatial and temporal embeddings, which are finally processed by fully connected layers to generate the final traffic inflow/outflow prediction, $\mathbf{N}^{\text{Taxi}}_{d,T}$, for the $T$-th time interval of the $d$-th day.

\subsection{Spatial and Temporal Embedding Modules} 
The role of this module is to process the traffic flow graph adjacency matrices and node features and produce spatial embeddings. Specifically, given adjacency matrix $\mathbf{G}^{\text{mode}}_{d,t}$, and node features $\mathbf{N}^{\text{mode}}_{d,t}$, $\text{mode} \in \{\text{Taxi}, \text{Lyft}\}$, we process them by graph convolutional network (\texttt{GCN}), followed by a \texttt{Dropout} layer \cite{goodfellow2016deep} for regularization and a fully connected layer (\texttt{FC}) to get the spatial node embeddings $\hat{\mathbf{N}}^{\text{mode}}_{d,t} \in \mathbb{R}^{2 \times M}$, \textit{i.e.,} 
\begin{equation}
    \hat{\mathbf{N}}^{\text{mode}}_{d,t} = \texttt{FC}\left(\texttt{Dropout}\left(\texttt{GCN($\mathbf{G}^\text{mode}_{d,t}$, $\mathbf{N}^\text{mode}_{d,t}$})\right)\right).
\end{equation}

The role of this module is to process the output of the spatial embedding module to produce spatial and temporal embeddings. In particular, given spatial taxi node embeddings in time interval $T-k$ to $T-1$, $\{\hat{\mathbf{N}}^\text{Taxi}_{d,T-k}, \dots, \hat{\mathbf{N}}^\text{Taxi}_{d,T-1}\}$ of the $d$-th day and the spatial Lyft node embeddings in $T$-th time interval of the $(d-1)$-th day, denoted as $\hat{\mathbf{N}}^\text{Lyft}_{d-1,T}$, we process them by an long short-term memory (\texttt{LSTM}) layer followed by three fully connected (\texttt{FC}) layers to produce the final traffic inflow/outflow prediction for, \textit{i.e.,}
\begin{equation}
    \mathbf{N}^{\text{Taxi}}_{d,T} = \texttt{FC}\left(\texttt{FC}\left(\texttt{FC}\left(\texttt{LSTM}\left(\hat{\mathbf{N}}^\text{Lyft}_{d-1,T}, \{\hat{\mathbf{N}}^\text{Taxi}_{d,T-k}, \dots, \hat{\mathbf{N}}^\text{Taxi}_{d,T-1}\}\right)\right)\right)\right).
\end{equation}

\section{Experimental Studies}
\label{sec:experiments}
\subsection{Experimental Settings}

$\bullet$ \textbf{Baselines:} In this prototype study, 
we compare our proposed method with the following deep learning-based baselines:

\begin{enumerate}[align = left]
    \item[(1)] \textbf{GCN}: We consider a model based on graph convolutional neural networks (GCN) as a baseline, which accepts the same input as our proposed framework.
    \item [(2) -- (4)] \textbf{LSTM}, \textbf{GRU}, and \textbf{RNN}: We implement neural networks based on recurrent long short-term memory (LSTM), gated recurrent unit (GRU) and simple recurrent neural network (RNN) modules to compare with our framework, which consists of one recurrent layer followed by fully connected (FC) layers. We input the node features to these baselines to predict the inflow/outflow of the target time interval $T$.
    \item[(5)] \textbf{GAT}: We implement a neural network based on graph convolutional layers with attention mechanism~\cite{gat} to generate the graph embeddings, which is followed by fully connected layers to generate the final inflow/outflow of the target time interval $T$.
    
    \item[(6)] \textbf{CGCN}: We implement a network based on Chebyshev convolutional layer~\cite{chebygcn} to produce the graph embeddings, which is similarly followed by fully connected layers to generate the final output.
\end{enumerate}

$\bullet$ \textbf{Parameter Settings:} We set $P = 8$ and divide each day into 8 equal 3-hour time intervals. Also, we set $k=3$ and predict the inflow/outflow of each region for the time interval $T$ using the historical data of the last three time intervals. We use 70\% of our dataset for training and the rest for evaluation. Also, we use the Adam optimizer \cite{goodfellow2016deep} with a learning rate of $0.001$ to train our framework and the baselines. Moreover, we use 32 and 16 neurons for the GCN and the fully connected layers of the graph embedding module. Additionally, we use an LSTM layer with 32 neurons for the temporal embedding module with a linear activation function. Furthermore, all our \name{} and the baselines all consist of three fully connected layers consisting of 32, 128, and $\text{2}\times\text{265}$ neurons to generate the output with the appropriate shape. For baselines (2)--(4), we use 32 neurons for the recurrent components. Finally, for baselines (5) and (6), we use 32 neurons for the graph neural layers to generate the node embeddings.

$\bullet$ \textbf{Performance Metrics:}
We use the mean absolute error (MAE) and mean squared error (MSE) as metrics 
to measure the difference between the predictions and ground-truths of traffic inflows and outflows.

\subsection{Experimental Results}
\label{sec:exp-results}

We present our experimental results as follows. 

\begin{figure}
    \centering
    \includegraphics[width = 1\columnwidth]{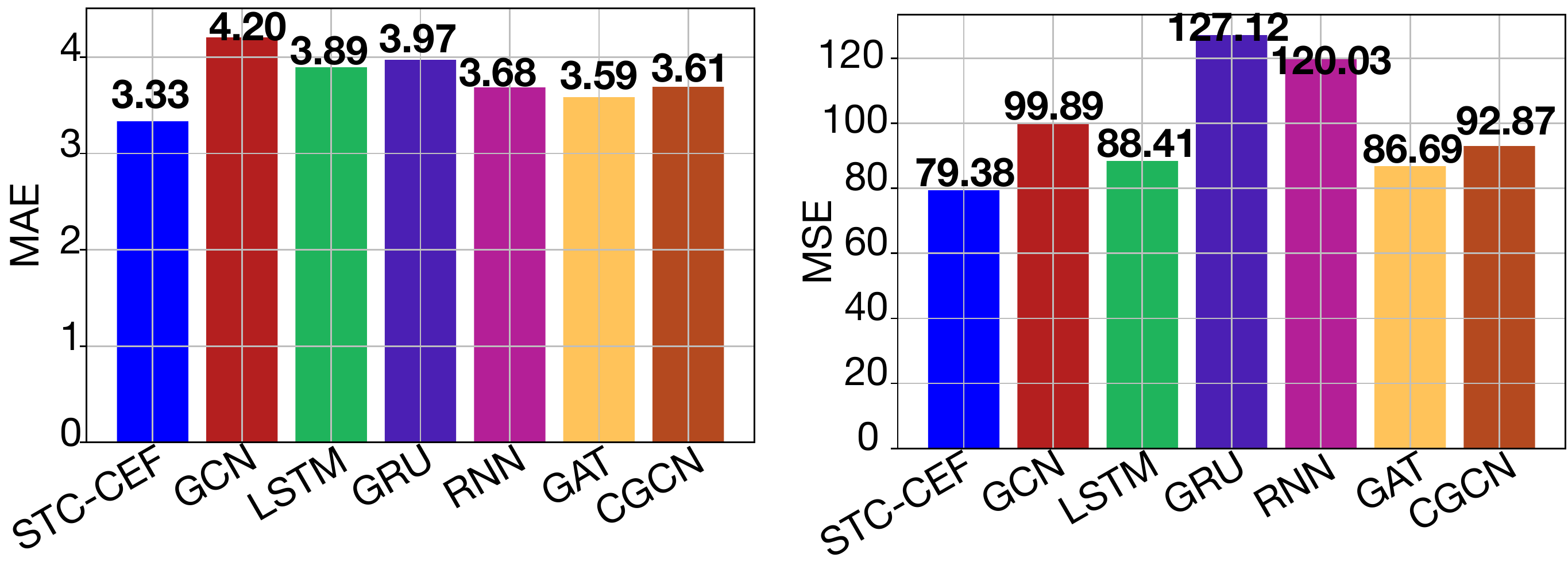}
    \caption{Overall performance (MAE and MSE) of \name{} and the baselines for traffic flow prediction.
    }
    \label{fig:overall}
\end{figure}

$\bullet$ \textbf{Overall Performance:}
We feed the historical taxi and Lyft trip data to our \name{} and the baselines for performance evaluation and present the overall results in Fig.~\ref{fig:overall}. We can observe that our \name{} outperforms the other baselines in terms of MSE and MAE errors. Furthermore, the figure implies that models based on graph neural networks perform better than the methods based on recurrent neural networks. However, the combination of both graph and recurrent neural layers is required for more effective traffic flow prediction.

\begin{figure}
    \centering
    \includegraphics[width = \columnwidth]{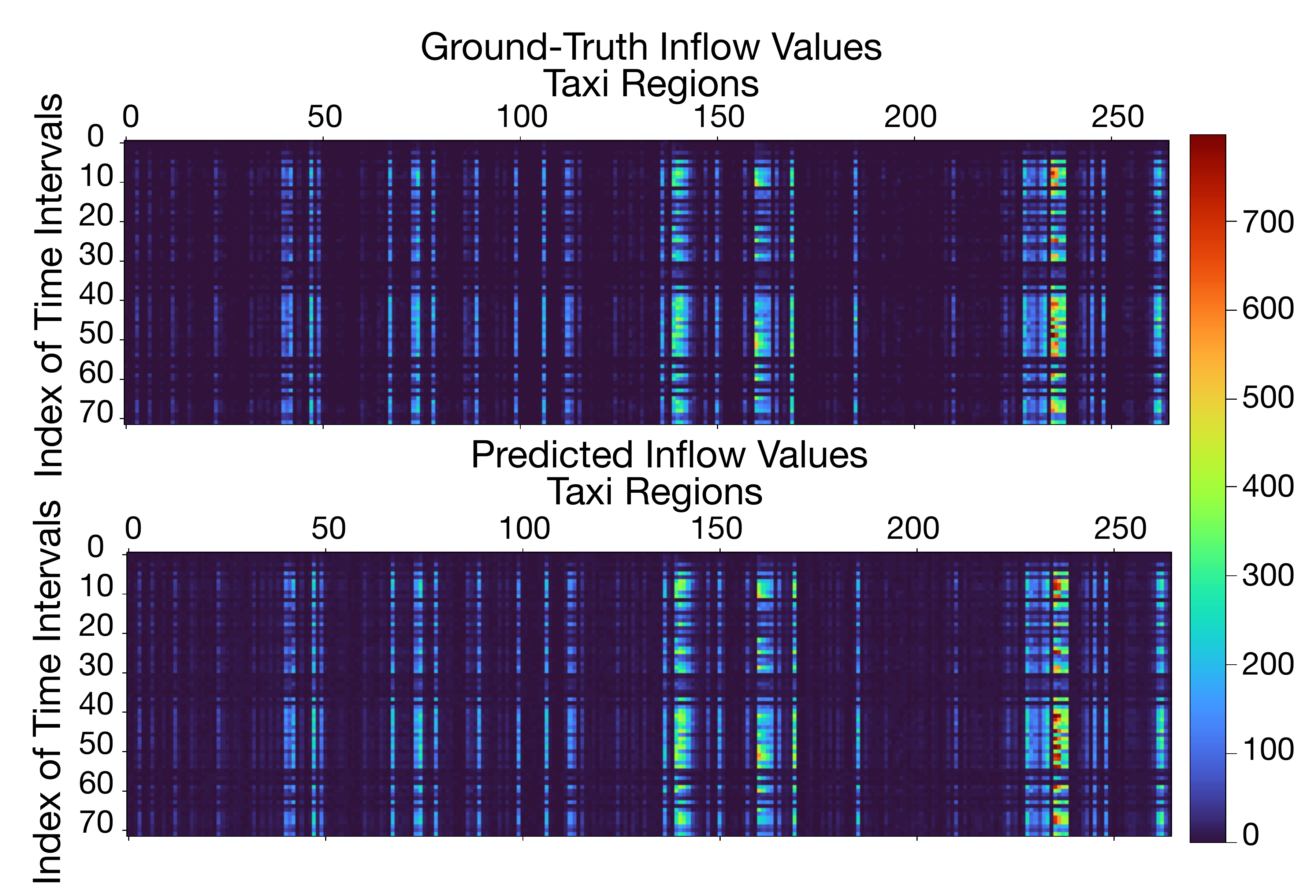}
    \caption{Illustration of  inflow prediction compared with the ground-truth values by \name{}.}
    \label{fig:pred-sample}
\end{figure}

\begin{figure}
    \centering
    \includegraphics[width =0.86\columnwidth]{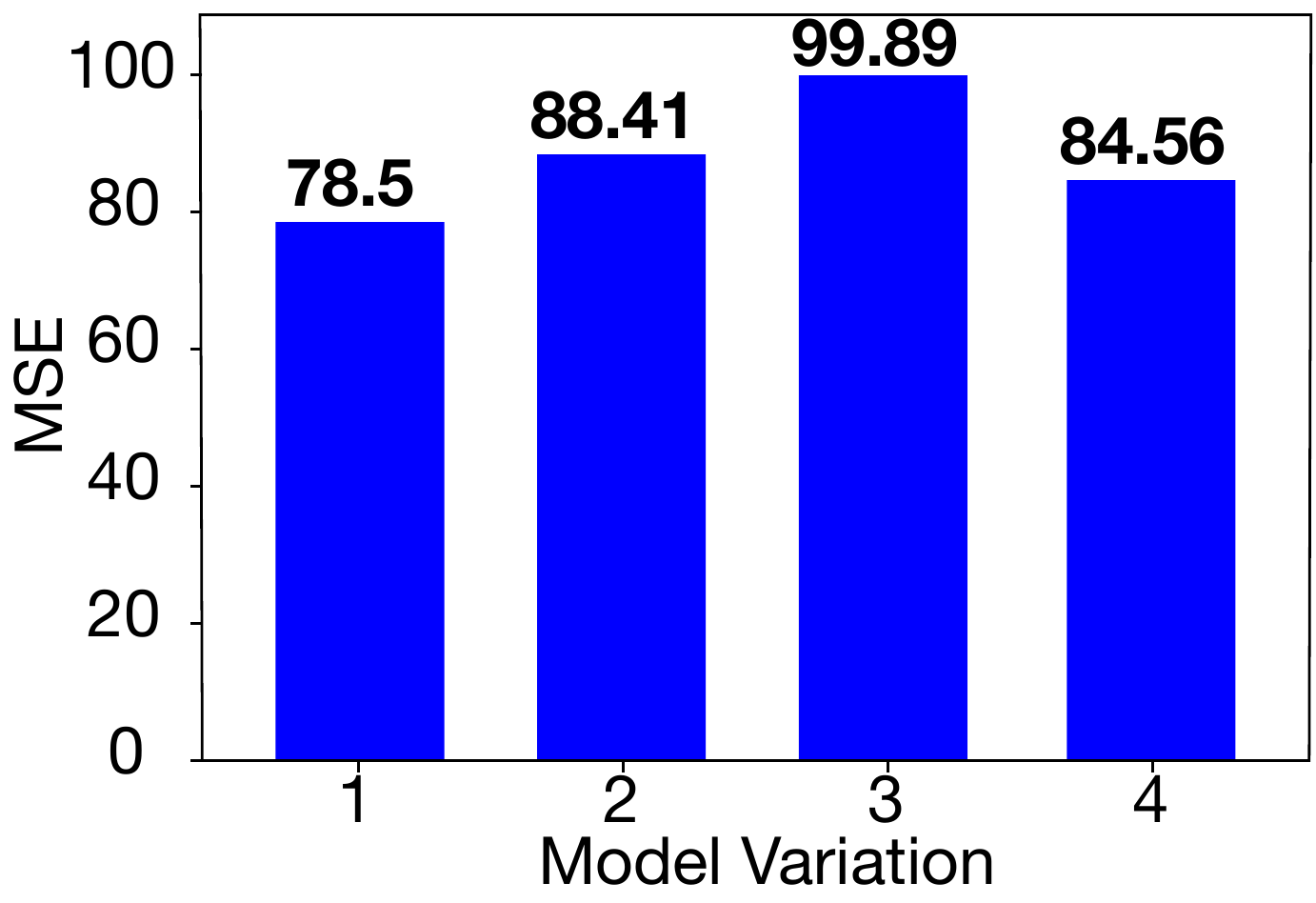}
    \vspace{-0.026in}
    \caption{Model ablation studies.}
    \label{fig:ablation}
\end{figure}

$\bullet$ \textbf{Ablation Studies:} To show the importance and role of each component in our framework, we perform ablation studies on different variations of our framework. 
In particular, we consider the following variations: 
\textit{(1)} the complete model, 
\textit{(2)} without the graph embedding module, 
\textit{(3)} without the temporal embedding module, 
and \textit{(4)} without fusing other transportation platform data (Lyft trip data in our studies). 
We illustrate the results in Fig.~\ref{fig:ablation}, from which we can see 
that the highest performance drop is caused by removing the temporal embedding module (variation 3). 
Furthermore, we can observe that removing the spatial embedding module (variation 2) also reduced the model's performance, thus showing the effectiveness of this module. Finally, we can see that by removing the Lyft trip data, the model's error is increased, which implies that the fusion of different transportation data is necessary for higher traffic flow prediction accuracy. 

$\bullet$ \textbf{Visualization:} We illustrate the predictions of our model compared to the ground truth traffic inflow values for the test data in Fig.~\ref{fig:pred-sample}, in which the warm colors indicate a high inflow/outflow rate between the corresponding regions while cooler colors represent otherwise. We can observe that our model's prediction is significantly close to the ground truth values. Additionally, by calculating the difference between the ground truth and the predicted values, we realized only 5\% of the regions had an error higher than the average error reported in Table.~\ref{fig:overall}, which further shows the effectiveness of our model.

\section{Conclusion}
\label{sec:conclusion}
We have proposed \name{}, a novel traffic prediction architecture 
based on graph convolutional and recurrent neural networks that 
extracts spatial and temporal information from the traffic flow 
data across different transportation platforms.
We have designed a novel spatio-temporal graph embedding approach 
to fuse the traffic flow data from taxis and ride-sharing platforms. 
We have performed experimental studies to show the effectiveness of 
\name{}, and our results have demonstrated 
that our model has outperformed the baseline models in traffic flow prediction.
In our future work, we will further conduct extensive data analytics for
more comprehensive model studies, including comparison with more state-of-the-art approaches.

\section{Acknowledgment}
\label{sec:ack}
This project is supported, in part, by the 2021 Google Research Scholar Program Award and the 2021
NVIDIA Applied Research Accelerator Program Award. 

\newpage

\bibliographystyle{ACM-Reference-Format}
\bibliography{references}

\end{document}